\documentclass{article}

\usepackage{times}
\usepackage{graphicx} 
\usepackage{subfigure}
\usepackage{caption}

\usepackage{natbib}

\usepackage{hyperref}

\usepackage[accepted]{template}

\icmltitlerunning{Power Benchmarks on Neuromorphic Hardware}

\begin{document} 

\twocolumn[
\icmltitle{Benchmarking Keyword Spotting Efficiency on Neuromorphic Hardware}

\begin{icmlauthorlist}
\icmlauthor{Peter Blouw}{}
\icmlauthor{Xuan Choo}{}
\icmlauthor{Eric Hunsberger}{}
\icmlauthor{Chris Eliasmith}{}
\item{Applied Brain Research, Inc.}
\item{Waterloo, ON, Canada}
\item{Correspondence: \{peter.blouw, xuan.choo, eric.hunsberger, chris.eliasmith\}@appliedbrainresearch.com}

\end{icmlauthorlist}

\icmlkeywords{keyword spotting; speech recognition; machine learning; neuromorphics; spiking neural networks}
\printAffiliationsAndNotice{\icmlEqualContribution} 

\vskip 0.3in
]

\begin{abstract} 

 Using Intel's Loihi neuromorphic research chip and ABR's Nengo Deep Learning toolkit, we analyze the inference speed, dynamic power consumption, and energy cost per inference of a two-layer neural network keyword spotter trained to recognize a single phrase. We perform comparative analyses of this keyword spotter running on more conventional hardware devices including a CPU, a GPU, Nvidia's Jetson TX1, and the Movidius Neural Compute Stick. Our results indicate that for this real-time inference application, Loihi outperforms all of these alternatives on an energy cost per inference basis while maintaining equivalent inference accuracy. Furthermore, an analysis of tradeoffs between network size, inference speed, and energy cost indicates that Loihi's comparative advantage over other low-power computing devices improves for larger networks.  

\end{abstract} 

\section{Introduction}

Neuromorphic hardware consists of large numbers of neuron-like processing elements that communicate in parallel via spikes. The use of neuromorphic devices can provide substantial improvements in power efficiency and latency across a wide range of computational tasks. These improvements are largely due to (a) an event-driven model of computation that results in processing elements only emitting spikes (and thereby consuming energy) when necessary, and (b) a substantial amount of architectural parallelism that reduces the number of computational `steps' required to transform an input into an output. 

\begin{figure}[ht!]
\centering
    \includegraphics[width=3.2in]{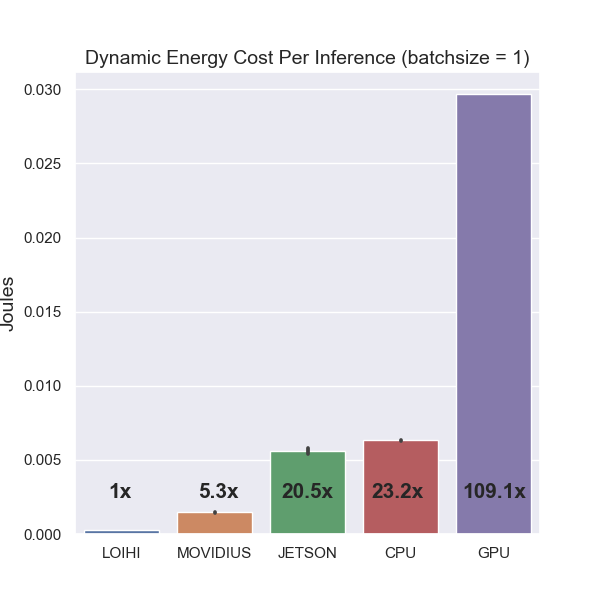}
    \caption{Dynamic energy cost per inference across hardware devices. This metric is equal to the difference between the total energy consumed by the hardware device over the time it takes to perform one inference minus the energy consumed over the same amount of time while the hardware is idling. For the keyword spotter, one inference involves passing a feature vector through a feed-forward neural network with two hidden layers to predict a probability distribution over alphabetical characters.}
\label{per_inf_fig}
\end{figure}

To provide a quantitative analysis of these expected benefits, we present the results of benchmarking a keyword spotting model running on Intel's new neuromorphic research processor, Loihi \cite{Davies:2018}. Keyword spotting involves monitoring a real-time audio stream for the purposes of detecting some keyword of interest (e.g. ``Hey Siri''). This task is useful for benchmarking neuromorphic devices because it (a) requires low-latency processing of real-time input signals, (b) benefits considerably from improvements in energy efficiency, and (c) has numerous practical applications in mobile and IoT devices.

\begin{table*}[t!]
\caption{Mean power consumption and energy cost per inference across hardware devices.} 
\vskip 0.15in
\begin{center}
\begin{small}
\begin{sc}
\begin{tabular}{lcccccc}
\hline
\abovespace\belowspace
Hardware & Idle (W) & Running (W) & Dynamic (W) & Inf/Sec & Joules/Inf \\
\hline
\abovespace
GPU & 14.97 & 37.83 & 22.86 & 770.39 & 0.0298 \\   
CPU & 17.01 & 28.48 & 11.47 & 1813.63 & 0.0063 \\
Jetson & 2.64 & 4.98 & 2.34 & 419 & 0.0056 \\
Movidius & 0.210 & 0.647 & 0.437 & 300 & 0.0015 \\
\belowspace
Loihi & 0.029 & 0.110 & 0.081 & 296 & 0.00027 \\

\hline
\end{tabular}
\end{sc}
\end{small}
\end{center}
\vskip -0.1in
\label{power-table}
\end{table*}

To investigate the relative power efficiency of different types of hardware running our keyword spotter, we perform experiments using the following devices: CPU (Xeon E5-2630), GPU (Quadro K4000), Jetson TX1, Movidius NCS, and Loihi (Wolf Mountain board). Below, the methodology for each experiment is reported, along with a detailed discussion of results. The same inference-only version of the keyword spotter running in TensorFlow (TF) is used in each non-spiking benchmark, achieving a test classification accuracy of 92.7\% across all hardware devices on a small dataset of spoken utterances we collected. A model that is architecturally identical to this TF model is then trained to work in the spiking domain \cite{Hunsberger:2016} and run with the Nengo neural simulator \cite{Bekolay:2014}, achieving a test classifcation accuracy of 93.8\% both in simulation and when running on Loihi.

Overall, our comparisons indicate that Loihi is more power efficient on a cost-per-inference basis than alternative hardware devices in the context of this keyword spotting application (see Figure \ref{per_inf_fig} and Table \ref{power-table}), and more generally in the context of similar deep learning applications.

In the remainder of this paper we describe the network and training methods in more detail, discuss our data collection methods, and present and discuss in-depth results. Code for reproducing our results is available at \url{https://github.com/abr/power\_benchmarks/}.

\section{Methodology}

The purpose of the keyword spotting network is to take in an audio waveform corresponding to an utterance, and then predict a sequence of characters to ascertain whether the utterance contains some keyword of interest. The audio waveform is preprocessed by performing Fourier transforms on an overlapping series of windows to compute Mel-frequency Cepstral Coefficient (MFCC) features for each window. Windows of adjacent MFCC features are concatenated into frames and passed as input to the network, with the frames having a stride of 10ms.  

\begin{table*}[t!]
\vskip 0.15in
\begin{center}
\begin{small}
\begin{sc}
\begin{tabular}{lcccccc}
\hline
\abovespace\belowspace
Model & True Positive (\%) & False Negative (\%) & True Negative (\%) & False Positive (\%) \\
\hline
\abovespace
TensorFlow & 92.7 & 7.3 & 97.9 & 2.1 \\   
\belowspace
Nengo Loihi & 93.8 & 6.2 & 97.9 & 2.1 \\

\hline
\end{tabular}
\end{sc}
\end{small}
\end{center}
\vskip -0.1in
\caption{Keyword spotting performance analysis. The spiking network on Loihi classifies examples as accurately as the non-spiking ANN in TensorFlow. The performance improvement seen on Loihi is not significant, since it amounts to the correct classification of a single additional test item.}
\label{acc-table}
\end{table*}

\begin{figure}[ht!]
\centering
    \includegraphics[width=2.5in]{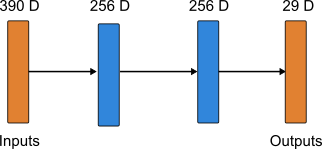}
    \caption{Network topology for the keyword spotter run on all devices. All implementations of the network make use of the same TensorFlow computational graph and parameters, with the exception of the Loihi implementation, which has the same network structure but is run in spiking mode with Nengo. On all devices, audio features are passed to the input layer and probability distributions over characters are read from the output layer, ensuring that identical amounts of computation are performed across all comparisons.}
\label{network_fig}
\end{figure}

Once a frame (390 dimensions) is provided as input to the network, it is passed through two 256-dimensional hidden layers and then used to predict a 29-dimensional output vector that corresponds to a probability distribution over alphabetical characters. Figure \ref{network_fig} provides a visual depiction of this network topology. Passing all frames derived from the initial audio waveform through the network yields a sequence of character predictions that gets collapsed by merging repeated characters and stripping out special characters corresponding to letter repetitions and silence. 

\begin{figure*}[ht!]
\centering
    \includegraphics[width=6.5in]{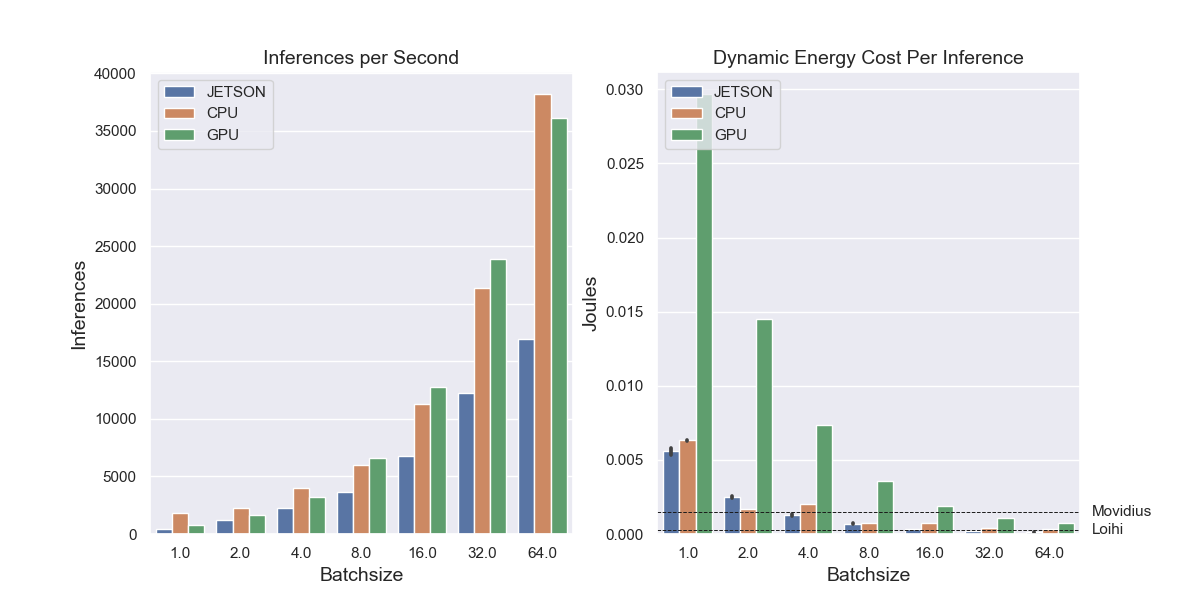}
    \caption{Inference speed and energy cost per inference as a function of batchsize. The horizontal lines indicate the energy cost per inference on Movidius and Loihi for a batchsize of 1. The intersection of the plot with these lines indicates the points at which batching reduces cost per inference to a level that is comparable with these architectures.}
\label{batchsize_fig}
\end{figure*}

We are interested in comparing the energy cost per inference for the keyword spotter on different kinds of hardware. To do this for all non-Loihi devices, we make use of a single script that runs a loop that performs a single forward pass of the model per iteration. The script does not change across trials; only the hardware targeted by the inference call in the loop does. We then run the loop for a specified duration of time (15 minutes) while a logging script is running in the background to write power readings to a CSV file every 200ms. A timestamp is saved upon entering and exiting the loop, and these time stamps are used to extract the corresponding range of power readings from the CSV file. We track the total number of inferences performed over the loop execution, and compute the average number of inferences performed per power reading (i.e., inferences per 200ms). By default, the batchsize is one, so a forward pass performs one inference; it is possible to adjust the batchsize, in which case the number inferences is calculated as the product of the number of loop iterations and the batchsize. As shown in Figure \ref{batchsize_fig}, batching can significantly improve power efficiency on a cost-per-inference basis, but can generally only be done in cases involving offline data analysis (since real-time data streams are typically unbatched). It is also possible to scale the internal architecture of the model (e.g., by increasing the number of neurons in each hidden layer), which enables keeping I/O constant while varying the compute load per loop iteration. Overall, with the average time per power reading and the average number of inferences per power reading, we can compute an estimate of the number of joules per inference for each power reading.

\subsection{Speech Data Collection and Model Training}

To get training data for the keyword spotter, we used Amazon's Mechanical Turk platform. We collected several thousand examples of (audio, text) pairs covering both a target phrase (``aloha'', in keeping with Loihi's Hawaiian theme) and close distractor phrases (e.g., ``take a load off''). After preprocessing the resulting dataset to screen out samples with poor audio quality or mismatches between the label phrase and the spoken utterance, we are left with a training set of approximately 2000 utterances from 96 speakers, split in a roughly 3:1 ratio between positive and negative examples of the target phrase. We seperate out a test set of 192 utterances consisting of one positive and one negative example of the target phrase per speaker. 
 
As mentioned, our basic neural network model consists of two 256-dimensional hidden layers, and is trained with the connectionist temporal classification (CTC) loss function \cite{Graves:2006} in TensorFlow \cite{Abadi:2016}. These design choices are in keeping with existing work on keyword spotting \cite{Chen:2014}, and with research on speech recognition more generally. Our trained TensorFlow model runs identically on all devices except Loihi, achieving a classification accuracy of 92.7\% on positive examples of the target phrase drawn from the test set. We use Nengo DL \citep{Rasmussen:2018} to learn a set of parameters for a spiking neural network with the same architecture as the non-spiking network that best approximates the performance of the original TensorFlow model. A comprehensive analysis of the performance of the spiking and non-spiking networks is provided in Table \ref{acc-table}, with all Nengo Loihi results obtained identically on both the Loihi chip itself and a software emulator of the chip. Note that given the small size of the test dataset (96 positive and negative examples), the difference between e.g., the true positive rates for the different models amounts to a very small difference in the total number of true positive classifications. Further evaluations of the functional characteristics of spiking networks on Loihi would likely benefit from tests involving larger datasets. 

\subsection{Power Measurements}

To account for idle power consumption on each device, we subtract a baseline average computed over 15 minutes to account for variability introduced by e.g., background OS processes that may run intermittently. With this baseline average in hand, we can compute dynamic power consumption for each hardware backend, and a corresponding estimate of dynamic joules per inference. More generally, for logging of power data, we run the model for 15 minutes to provide a good estimate of the true distribution of power readings that can be expected for a given hardware device.

In the case of Loihi, we use a separate script that prepares the inputs in accordance with the requirements of Intel's API for Loihi (NxSDK), such that a collection of input spikes are injected into Loihi's neuromorphic mesh. Once these input spikes are presented to perform inference on the chip, we can directly measure power consumption over the course of runtime and thus obtain the timestep per power reading, the number of inferences per timestep, and the number joules per inference for each reading. We use runtimes of approximately 15 seconds on Loihi.

\begin{figure*}[ht!]
\centering
    \includegraphics[width=6.5in]{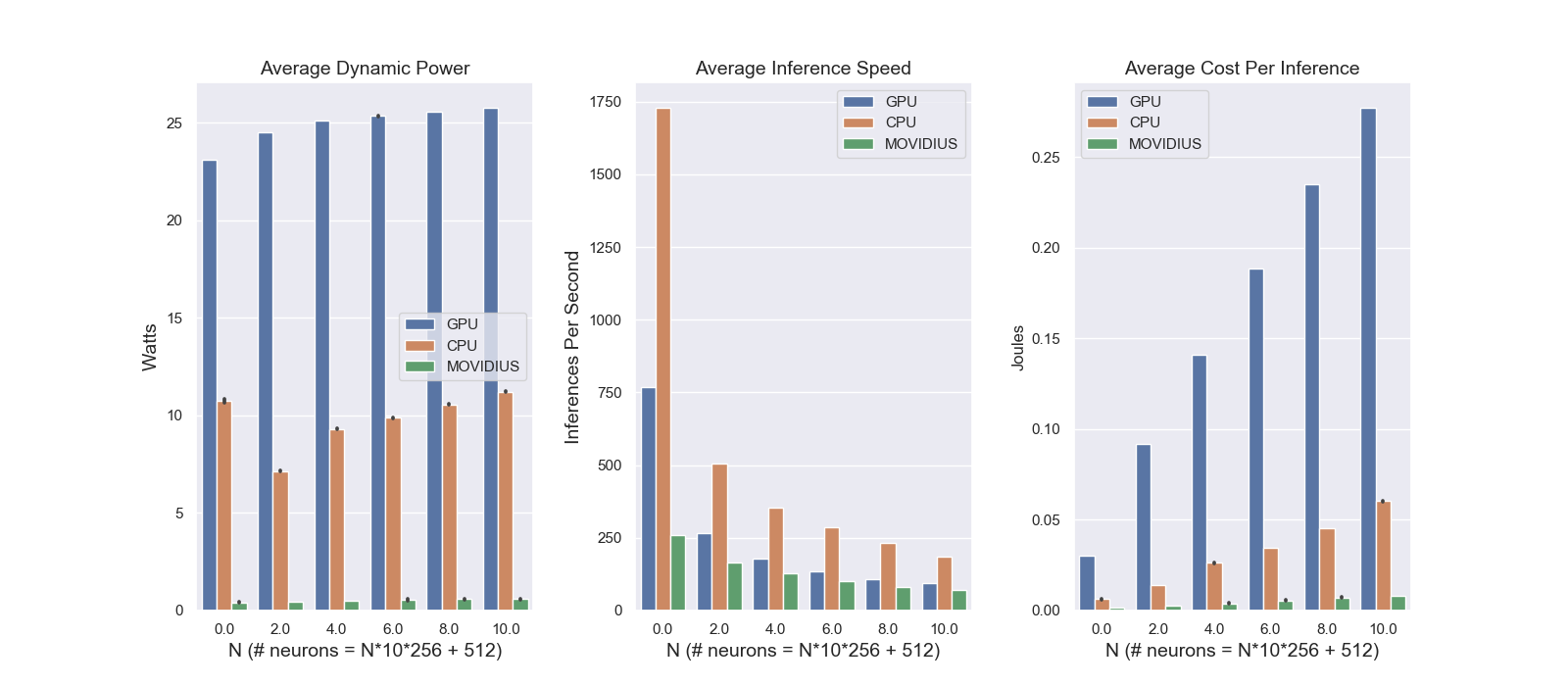}
	\caption{A comparison between GPU, CPU, and Movidius NCS for inference speed and energy cost per inference as a function of network size. The specific scaling factor is chosen to enable identical linear increases in network size across all devices. Comparisons to Loihi are included in a seperate plot below to aid visualization. A fixed uniform distribution is used to randomly generate weights on all devices, so the networks examined here are not identical to the keyword spotter. The network topology is presented in Figure \ref{scaled_network_fig} below.}
\label{comp_fig}
\end{figure*}

In brief, our methodology is as follows: (1) estimate idle power consumption; (2) log power readings during runtime; (3) estimate dynamic power by subtracting idle baseline from log; (4) estimate average logging interval and average number of inferences per interval; and (5) estimate dynamic joules per inference from (3) and (4).

Importantly, it is currently not possible to use automated logging for the Jetson and Movidius platforms, which reduces the precisions of our estimates of the power readings for idling and runtime. The number of inferences performed over the course of runtime can still be computed exactly though, so we can nonetheless estimate the dynamic energy cost per inference for these latter platforms. For all reported results, the batchsize is 1 unless otherwise specified.

To ensure clarity regarding the content of our results, there are a number of assumptions built into our experiments that are outlined in the following sections.

\subsubsection{Software Variability}

\begin{figure*}[ht!]
\centering
    \includegraphics[width=6.7in]{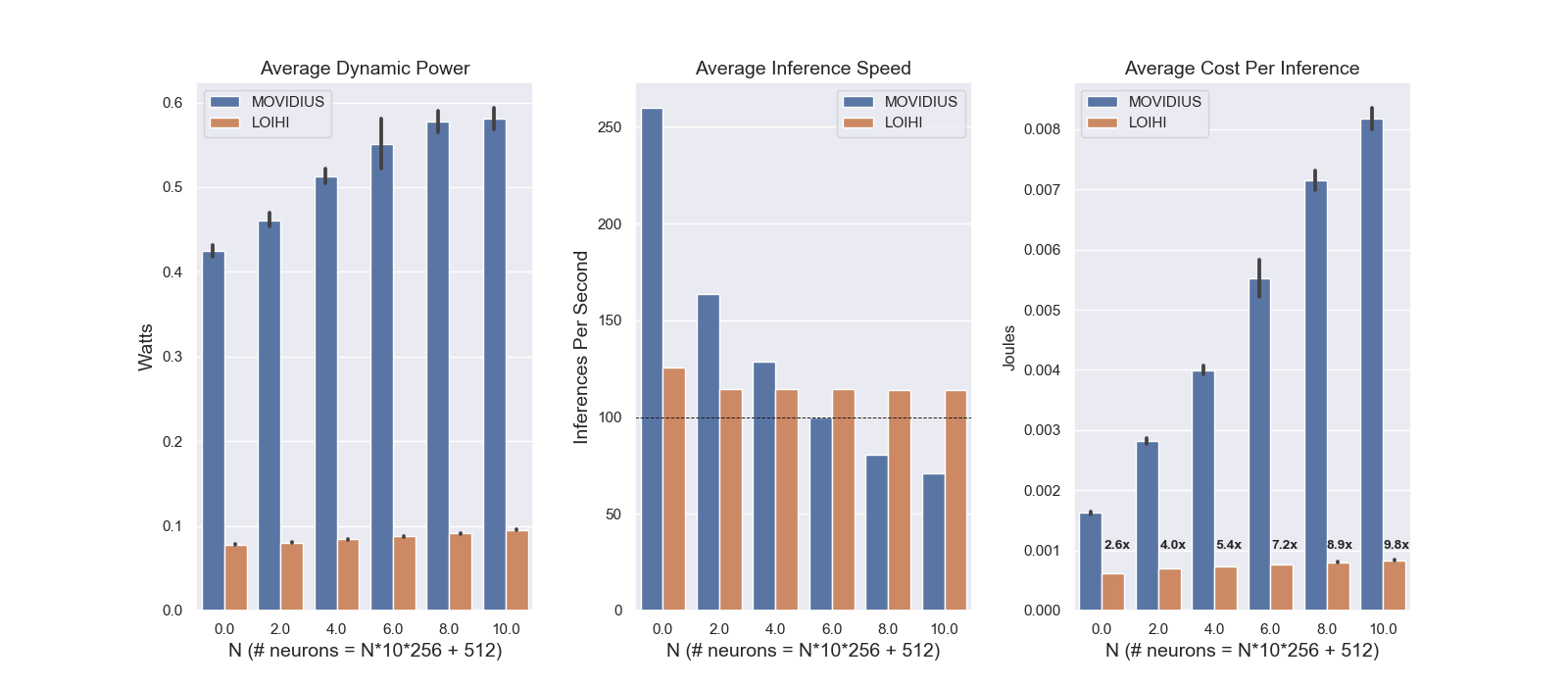}
	\caption{A comparison between Movidius NCS and Loihi for inference speed and energy cost per inference as a function of network size. The scaling architecture and parameter initializations are identical to those used in Figure \ref{comp_fig} above. The dotted line in the middle plot indicates the speed threshold for implementing real-time inference on audio data with a frame stride of 10ms. With larger network sizes, Loihi maintains its ability to perform real-time inference, while the Movidius NCS does not.}
\label{scaling_fig}
\end{figure*}

During runtime, inference execution involves a complicated software stack that is non-identical across hardware backends. For example, on CPU, GPU, and Jetson, inference is performed directly in TensorFlow, which controls how each computation is executed and how data transfer is performed. On Movidius and Loihi, by comparison, custom APIs are used to perform these tasks using hardware-specific compilations of the original TensorFlow model. As such, differences between these software stacks are incorporated into our power measurements, and it is generally not possible to disentangle changes in power consumption that are due to the software stack from changes due to the hardware itself. We have let TensorFlow allocate memory, perform data I/O, and execute computations as per default, which is likely to be comparable to how a typical TensorFlow user would try to build and deploy a model of this sort. In the case of GPU computations, this may provide a distorted view of energy consumption given that TensorFlow tries to allocate 100\% of GPU memory by default, regardless of the computation graph being implemented (we have set the per process allocation limit to 10\% of available memory, though this does not seem to significantly reduce dynamic power draw). Further differences could theoretically arise if the experiments are repeated with different versions of e.g., TensorFlow, CUDA, CuDNN, or any other packages present in a particular software stack. In the case of CPU benchmarks, our benchmarking workstation includes 24 Xeon E5-2630s, and we allow TensorFlow to allocate computations across these cores as per its default settings. 


\subsubsection{Data I/O}

\begin{figure*}[ht!]
\centering
    \includegraphics[width=6.7in]{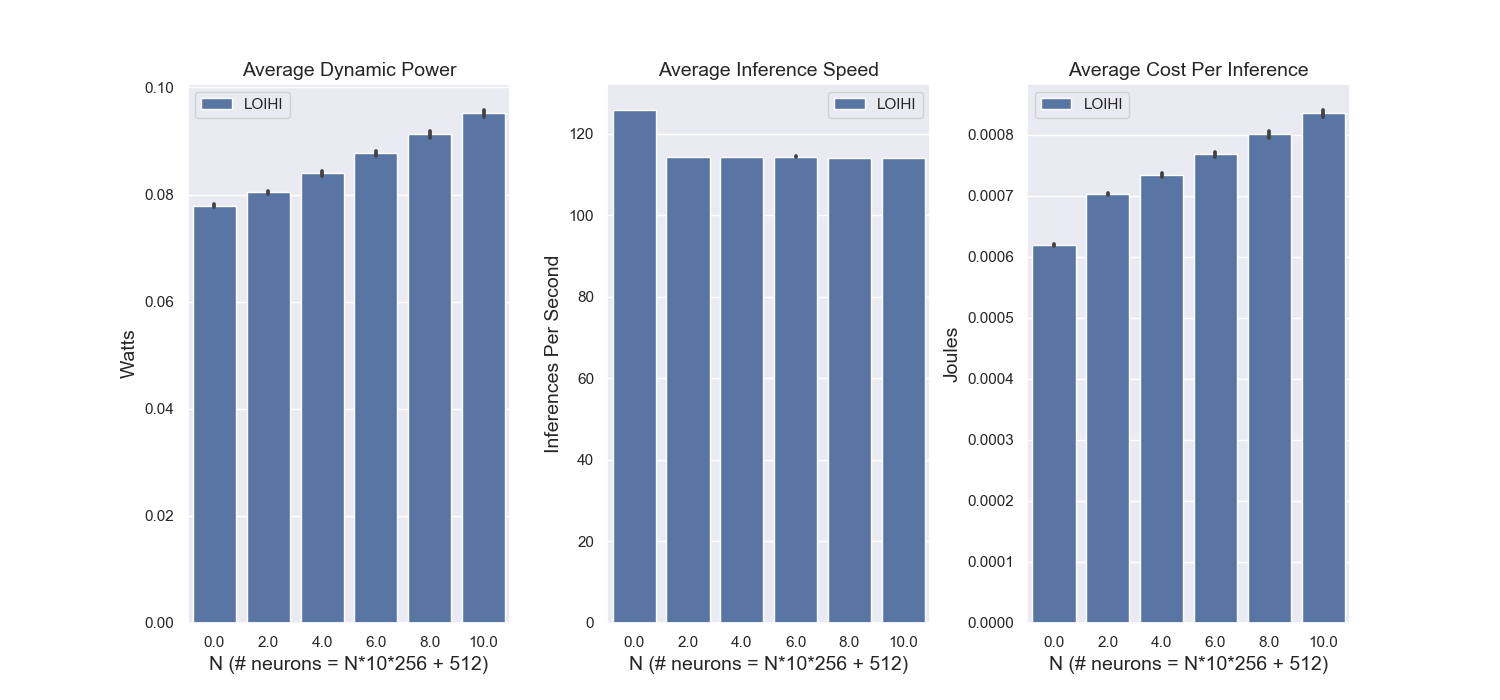}
    \caption{A closer analysis of system performance as a function of network size on Loihi. The roughly linear increase in dynamic energy consumption and cost per inference is very modest as a function of network size. This leads to an increasingly large comparative advantage over Movidius as network size grows.}
\label{loihi_fig}
\end{figure*}

For every kind of hardware except for Loihi, we cannot directly characterize energy spent on I/O in relation to energy spent on compute. This is important because in the case of the Movidius in particular, it is possible that a significant proportion of the energy consumption is due to I/O, which could in turn exaggerate the difference between Movidius and Loihi. To help isolate this factor, we have evaluated different sized networks all with constant I/O. Hence, the constant energy associated with I/O is always bounded by the smallest network size evaluated (i.e., when N=0 in Figure \ref{scaling_fig}).

It is worth emphasizing here that we have adopted an extremely conservative approach to measuring dynamic power consuption on all devices that makes it unlikely that I/O differences are responsible for Loihi's observed efficiency gains. Specifically, we use the difference between runtime energy consumption and the idle power consumption measured after system boot (as opposed to idle power consumption measured after the model has completed execution at least once). In the case of Loihi, the difference between idling from boot and idling after model execution is approximately 70mW, which in proportional terms is substantially higher than the difference on other devices. We speculate that the embedded x86 processors used to manage the interface to Loihi's neuromorphic mesh are only activated upon executing a model for the first time, after which they return to an idling state. If so, then our measurements of dynamic power consumption on Loihi are overestimated such that it is highly unlikely for the inclusion of I/O power on the Movidius NCS to eliminate Loihi's advantage from an efficiency perspective. 

\subsubsection{Measurement Precision}

In the case of GPU, CPU, and Loihi, we have automated logging tools that can be used to record power measurements with a reasonable degree of temporal resolution and consistency. In the case of the Movidius and Jetson, we are forced to use a visual monitoring device, which reduces precision. In all cases, we log the displayed voltage and amperage every 15 seconds during each 15min inference runtime. A further point concerning Jetson is that we are forced to treat the combination of the CPU and GPU as the targeted hardware device, since we do not separately measure the power consumed by these two processors. A similiar point applies to Loihi and Movidius, since all of these devices contain embedded CPUs that interface with other accelerators to evaluate the output of the neural network.

There are also five sources of power consumption on Loihi that are incorporated into our measurements: (1) silicon leakage while the system is powered; (2) embedded x86 CPU idling power consumption; (3) core barrier sync power consumption; (4) spike processing power consumption during inference; and (5) embedded x86 CPU instructions involved in its interactions with the neuromorphic mesh on each timestep. Our measurements of idle power consumption for Loihi are intended include (1) and (2), while our measurements of runtime power consumption are additionally intended to include (3), (4), and (5). Note, as mentioned previously, that we are being very conservative in how we estimate dynamic power on Loihi (i.e., an argument could be made that we should not be including the 70mW overhead).

\section{Results} 
 
Table \ref{power-table} reports the mean idling, runtime, and dynamic power consumption for each device running the keyword spotter, along with the resulting mean inference speed and energy cost per inference. We use this data on power consumption and inference speed to compute the energy cost per inference as reported in Figure \ref{per_inf_fig}.

\begin{figure*}[ht!]
\centering
    \includegraphics[width=4in]{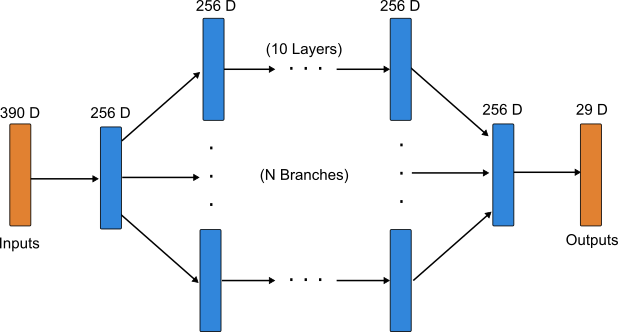}
    \caption{Topology for the network used in the scaling analyses reported in Figures \ref{comp_fig}, \ref{scaling_fig} and \ref{loihi_fig}. This topology is used because it is compatible with identical networks being run on all devices that scale to large numbers of neurons. Note that the network used in these scaling experiments is not the same as the original keyword spotter, even though in the special case of N=0, it has the same structure as the keyword spotter.}
\label{scaled_network_fig}
\end{figure*}

To explain, the mean idle power consumption is subtracted from the mean runtime power consumption to give the mean dynamic power consumption in Watts in the fourth column. Since one Watt is equivalent to one joule per second, we can divide the dynamic power consumption value by the number of inferences per second, listed in the fifth column, to get the mean energy cost per inference, listed in the final column. Bootstrapped 95\% confidence intervals (CIs) are depicted in all plots that make use of averages of multiple, distinct log values (e.g., Figure \ref{per_inf_fig}), although CI's with very tight bounds are in some cases not directly visible (e.g., the middle plot in Figure \ref{loihi_fig}). To compute CI's on the energy cost per inference, we subtract the mean idle power consumption for a given hardware device from all runtime log entries to convert the entries to dynamic power. Then, by calculating the average number of inferences per logging interval, we can compute the energy cost per inference for each logged power reading. This then lets us compute confidence intervals on the resulting average dynamic energy cost per inference on each device.

Since the energy cost per inference is highest for CPU and GPU, it is useful to look at how this energy cost changes as we modify the batchsize during inference. In the case of the GPU in particular, increasing the batchsize should increase energy efficiency by reducing the need to move data on and off the GPU before and after each individual inference. Figure \ref{batchsize_fig} indicates that, as one would expect, energy efficiency increases with batchsize, since all of the compute associated with feeding inputs and collecting outputs is only performed once per batch instead of once per inference. The flat lines on the graph indicate the energy cost per inference for Movidius and Loihi, so as to indicate the approximate batchsize at which generic hardware becomes comparable to these low-power alternatives from an efficiency perspective. Notably, both of these platforms only support a batch size of 1. In the case of Loihi this is because deployment is geared to real-time, interactive applications.

Figure \ref{comp_fig} displays the results of an experiment in which additional hidden layers are added to the model in accordance with the network topology in Figure \ref{scaled_network_fig}, but the batchsize is fixed at one. This keeps I/O fixed while varying the amount of computation required to perform an inference. Note, however, that since we cannot performance inference using our trained model parameters (which are for a different network structure), we are concerned only with how inference speed and cost-per-inference scale with model size. In other words, we are ignoring functionality (i.e., the network's weight values) and focusing only on tradeoffs between inference speed, inference cost, and network size. To a provide a linear increase in computation as a function of our independent variable N, we scale the network by adding multiple branches of multiple layers between the first and second hidden layers in the original model architecture, as shown in Figure \ref{scaled_network_fig}. As expected, inference speed decreases monotonically for all devices, while energy cost per inference increases monotonically. Dynamic power consumption also generally increases monotonically, except in the case of the CPU results. We speculate that the observed variability in dynamic power here has to do with how computations are allocated across cores as the ntework size (and corresponding compute load) increases.

Since Movidius is closest to Loihi from an efficiency perspective, it is useful to provide a more detailed analysis of how energy consumption and inference speed change as a function of network size in the two devices. Figure \ref{scaling_fig} illustrates how dynamic energy consumption, inference speed, and energy cost per inference change as a function of network size. Although the dynamic power consumption and cost per inference for Loihi appears to be flat in comparison to Movidius, Figure \ref{loihi_fig} indicates that, as expected, both metrics increase monotonically with network size. Moreover, Loihi's modest rate of increase in comparison to Movidius indicates a sizeable scaling advantage. This advantage likely arises both because Loihi's inherent architectural parallelism allows inference speed to remain roughly constant as a function of network size, and because the temporal sparsity and data locality of a spiking network leads total power consumption to increase much more slowly as function of network size. Together, these two factors enable Loihi to provide increasingly large performance and energy efficiency improvements over Movidius as network size grows. 

From a practical perspective, these gains are of considerable significance. Running real-time applications on audio data with a frame stride of 10ms requires an inference speed of at least 100 inferences per second. As is illustrated in Figure \ref{scaling_fig}, Loihi maintains this inference speed as network size increases, while Movidius does not. Since any form of speech recognition that goes beyond the keyword level will likely require a much more complicated network architecture, the performance characteristics of Movidius are likely prohibitive for such a task. In other words, systems like Loihi may offer the only practical path forward for building sophisticated programmable recognition systems on edge devices with constrained power and latency requirements.

It is also worth noting that although the network topology here is identical to that of the keyword spotter when N=0, the inference speed is a bit slower due to the fact that we allocate the neurons in the input layer across eight neuromorphic cores on Loihi (as opposed to two cores in the original keyword spotter) in order to accommodate the connectivity demands that arise for large values of N. This leads to an increase in the energy cost per inference for N=0 in comparison to the keyword spotter even though the number of neurons and weights remains the same. 

\section{Conclusion}

Overall, we have demonstrated that running a keyword spotting application on a state-of-the-art neuromorphic processor results in improved energy efficiency on a cost-per-inference basis over a number of existing low power computing devices. Given the results depicted in Figure \ref{scaling_fig}, it is worth noting that even if all of the dynamic power consumption for the Movidius NCS is spent on USB I/O when N=0, it would still be the case that Loihi offers a roughly 10x improvement in power efficiency when N=10, largely because Loihi's dynamic power consumption increases very modestly while its inference speed remains nearly constant due to the benefits of architectural parallelism. 

Overall, these scaling properties translate into significant energy efficiency improvements for large networks in comparsion to the closest alternative low-power device we tested, the Movidius NCS. These efficiency gains importantly come with no cost to model accuracy. Given as much, efforts to program neuromorphic devices to perform more sophisticated computational tasks are likely to yield significant benefits. 

\section*{Acknowledgements}

We thank Trevor Bekolay, Terry Stewart, and Dan Rasmussen for helpful suggestions regarding the software used to run the keyword spotter across all devices.

\section*{Code and Data} 
 
Code, data, and instructions for reproducing the results in this technical report are available at \url{https://github.com/abr/power\_benchmarks/}.

\bibliography{power_summary}
\bibliographystyle{template}

\end{document}